# Inference Using Message Propagation and Topology Transformation in Vector Gaussian Continuous Networks

**Satnam Alag**
Department of Mechanical Engineering
University of California at Berkeley
*alag@pawn.berkeley.edu*

**Alice M. Agogino**
Department of Mechanical Engineering
University of California at Berkeley
*aagogino@euler.me.berkeley.edu*

## Abstract

We extend continuous Gaussian networks – directed acyclic graphs that encode probabilistic relationships between variables – to its vector form. Vector Gaussian continuous networks consist of composite nodes representing multivariables, that take continuous values. These vector or composite nodes can represent correlations between parents, as opposed to conventional univariate nodes. We derive rules for inference in these networks based on two methods: message propagation and topology transformation. These two approaches lead to the development of algorithms, that can be implemented in either a centralized or a decentralized manner. The domain of application of these networks are monitoring and estimation problems. This new representation along with the rules for inference developed here can be used to derive current Bayesian algorithms such as the Kalman filter, and provide a rich foundation to develop new algorithms. We illustrate this process by deriving the decentralized form of the Kalman filter. This work unifies concepts from artificial intelligence and modern control theory.

## 1 Introduction

Monitoring and estimation problems involve estimating the state of a dynamical system using the information provided by the sensors. Estimation is the process of selecting a point from a continuous space – the best estimate or the expected value. In predicting, controlling, and estimating the state of a system it is nearly impossible to avoid some degree of uncertainty (Dean and Wellman, 1991). Probabilistic (Bayesian or belief) networks – directed acyclic graphs that encode probabilistic information between variables – have been found to be useful in modeling uncertainty. Our work aims at representing this estimation process by means of graphical structures and deriving rules for inference within these networks. By using this approach one can convert the tasks associated with monitoring and estimation: sensor validation, sensor data-fusion, and fault detection, to a problem in decision theory. The main advantage of this graphical approach is the ease with which old Bayesian algorithms such as the Kalman filter, the probabilistic data association filter, multiple model algorithms, as well as new algorithms can be represented and derived using these networks and rules for inference (Alag, 1996).

Practical systems consist of variables that acquire continuous values in an operating range, e.g., the temperature of a particular component being monitored or the pressure in a chamber. Therefore, to represent these variables by means of a network structure, the nodes that are associated with these variables should be able to take continuous values. Pearl (1988) has developed an encoding scheme for representing continuous variables in a belief network, which could potentially be used for our graphical framework. However, this representation, is based on the assumption that the parents of each variable are uncorrelated. This assumption breaks down when the underlying network contains loops, when two or more nodes posses both common descendants and common ancestors, which unfortunately could occur in the class of problems that we consider. We therefore develop a different encoding scheme for representing continuous variables in a belief network. This consists of extending continuous Gaussian networks to its vector form, where each node consists of a multivariate Gaussian distribution, as opposed to each node representing a univariate Gaussian distribution. These vector or composite nodes are capable of representing dynamic systems, where parent nodes may be correlated. These vector nodes correspond to the process of *clustering*: a method used to handle multiply connected networks. The links between these vector nodes are now characterized by matrices.

To apply these compound networks, we need rules for inference within these graphical structures. In this paper, we carry out this task for singly connected vector Gaussian networks. The case of multiply connected networks and mixed discrete-continuous vector Gaussian networks can be found in Alag (1996, Chapter 4) and is not addressed in this paper. We will develop rules for inference using first, the method of message propagation (Pearl, 1988; Driver and Morrell, 1995) and second, the method of topology transformation (Olmsted, 1984; Shachter, 1986; Rege and Agogino, 1988). These two methods have been chosen because they lead to the development of algorithms that can be implemented either in a decentralized (parallel using multiple processors) or a centralized (single processor) architecture.

In Section 2, we briefly review the previous work done in the area of continuous Gaussian networks and introduce vector Gaussian networks. In Section 3 we derive rules for inference using the method of message propagation, while in Section 4 we develop rules for topology transformation. We illustrate the potential of these networks by deriving the Kalman filter in Section 5 and present concluding remarks in Section 6. The appendix contains rules for multivariate normal distribution that were used in Section 3, as well as an alternate form of message to parents.

## 2 Continuous Gaussian Belief Networks

Consider a domain $\vec{x}$, of $n$ continuous variables $x_1,...,x_n$. The joint probability density function for $\vec{x}$ is a multivariate nonsingular normal distribution

$$\Pr(\vec{x}) = N(x; P, \bar{x}) \triangleq |2\pi P|^{-1/2} \exp\left(-\frac{1}{2}(x-\bar{x})^T P^{-1}(x-\bar{x})\right) \quad (2-1)$$

where $N(\cdot)$ denotes the normal pdf with argument $x$, and

$$\bar{x} = E[x] \text{ and } P = E\left[(x-\bar{x})^T(x-\bar{x})\right]$$

are respectively, the mean and covariance matrix. Throughout the paper we will use the symbol $Pr$ to represent probability, and the symbol $N()$ to represent the normal distribution. The inverse of the covariance matrix, $P^{-1}$ is also known as the precision or the information matrix.

This joint distribution of the random variables can also be written as a product of conditional distributions each being an independent normal distribution[1], namely

$$\Pr(x) = \prod_{i=1}^{n} \Pr(x_i | x_1,...,x_{i-1})$$

$$\Pr(x_i | x_1,...,x_{i-1}) = N\left(x_i, m_i + \sum_{j=1}^{i-1} b_{ij}(x_j - m_j), 1/v_i\right) \quad (2-2)$$

where $m_i$ is the unconditional mean of $x_i$, $v_i$ is the conditional variance of $x_i$ given values for $x_1,...,x_{i-1}$, and $b_{ij}$ is a linear coefficient reflecting the strength of the relationship between $x_i$ and $x_j$. Hence, one can interpret a multivariate normal distribution as a belief network, where there is an arc from $x_j$ to $x_i$ whenever $b_{ij} \neq 0$, $j < i$. This special form of belief network is commonly known as a Gaussian belief network. Readers are referred to Geiger and Heckerman (1994) for a more formal definition of Gaussian belief networks.

Pearl (1988, Section 7.2) has developed an encoding scheme for representing continuous variables in a belief network, which is based on the following assumptions[2] (1) all interaction between variables are linear (2) the sources of uncertainty are normally distributed and are uncorrelated (3) the belief network is singly connected[3].

Pearl (1988) considers a hierarchical system of continuos random variables, like the one shown in Figure 1. Each variable $X$ has a set of parent variables $U_1, U_2,..., U_n$ and a set of children variables $Y_1, Y_2,...Y_m$. The relation between $X$ and its parents is given by the linear equation

$$X = b_1 U_1 + b_2 U_2 + \cdots + b_n U_n + v \quad (2-3)$$

where $b_1, b_2,..., b_n$ are constant coefficients representing the relative contribution made by each of the $U$ variables to the determination of the dependent variable $X$ and $v$ is a noise term summarizing other factors affecting $X$. $v$ is assumed to be normally distributed with a zero mean, and uncorrelated with any other noise variables.

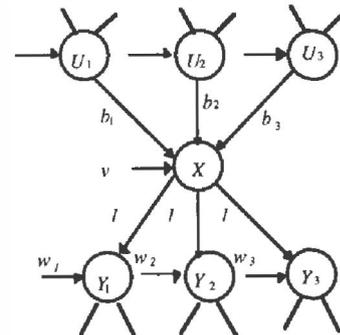

Figure 1: A fragment of a singly-connected network showing the relationships between a continuous variable $X$, its parent variables $U_1, U_2,..., U_n$ and the noise $v$.

Given the network topology, the link coefficients ($b$'s), the variances ($\sigma_v^2$) of the noise terms, and the means and variances of the root variables – nodes with no parents – Pearl (1988) has developed a distributed scheme for updating the means and variances of every variable in the network to account for evidential data $e$ – a set of variables whose values have been determined precisely. The update can also be done in a manner similar to that in Shachter and Kenley (1989). However, Pearl places an additional restriction that the computation be conducted in a *distributed* fashion, as though each variable were managed by a separate and remote processor communicating only with processors that are adjacent to it in the network, i.e., the update is performed in a decentralized (parallel) manner.

---

[1] This presentation is similar to that of Geiger and Heckerman, 1994.

[2] The first two assumptions are the same as those made by Kenley and Shachter, 1989.

[3] A network in which no more than one path exists between two nodes

The equation relating the variable of interest to its parents replaces the conditional probability table that is required in the case where the variables are discrete. In addition, due to the assumption that the variables are normally distributed the complete distribution can be specified with the help of just two parameters: the mean and the variance.

We extend these continuous Gaussian networks to their vector form. Figure 2 shows a generic form of a vector Gaussian network. Here, the variables $(U_i, X, Y_i)$ represented by the nodes are vectors, for e.g., $X \equiv [x_1, \cdots, x_n]^T$, where $x_i$ are Gaussian random variables.

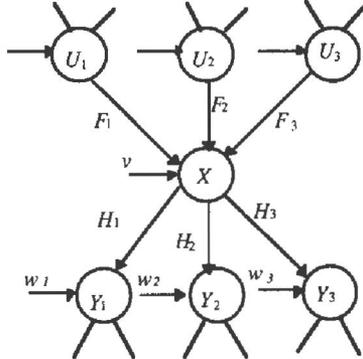

Figure 2: *Generic form of a vector Gaussian belief network.*

The arc between the variables represent the following relationship
$$X = \sum_i F_i \cdot U_i + v; \quad Y_i = H_i X + w_i \quad (2-4)$$
where $U_i$ and $Y_i$ are $n_{U_i}$ and $n_{Y_i}$ dimension vector, $F_i$ is a $n_x \times n_{U_i}$ matrix, $H_i$ is a $n_{Y_i} \times n_x$ matrix, $v$ is a $n_x$ vector, $Q = E[vv^T]$, the covariance matrix for the noise term represents the correlation between the parent variables, $w_i$ is a $n_{Y_i}$ vector. It can be easily shown that in the case of multiple parent nodes $[U_1, U_2]^T$ with corresponding state matrices $F_1, F_2$ respectively, one can obtain an equivalent matrix
$$F_{equivalent} = \begin{bmatrix} F_1 & 0 \\ 0 & F_2 \end{bmatrix}$$
which express the augmented state vector $[X_1, X_2]^T$. This procedure is known as clustering of the nodes. In general one may cluster the variables into nodes in which there is no evidence, nodes with different covariance matrices, correlated variables, etc. Each clustered (vector) node can be considered as made up of another Gaussian belief network where the inter-relationship between the variables of this continuous Gaussian belief network is given by the node's covariance matrix.

## 3 Inference using Message Propagation: Decentralized Approach

Consider a typical fragment of a singly connected network (Figure 2), consisting of an arbitrary node $X$, the set of all $X$'s parents, $U = \{U_1, U_2, \cdots, U_n\}$, and the set of all $X$'s children, $Y = \{Y_1, Y_2, \cdots, Y_m\}$. Let $e$ be the total evidence obtained, $e_x^-$ be the evidence connected to $X$ through its children ($Y$), and $e_x^+$ be the evidence connected to $X$ through its parents ($U$). The readers are referred to Pearl (1988; Section 7.2) where the rules for propagation are derived for the scalar case as shown in the network in Figure 1. To derive the inference rules for vector Gaussian networks, we will follow a procedure similar to Pearl's for the scalar case. Links in Pearl's network correspond to
$$x = \sum_i b_i u_i + v$$
$$y = x + w_i \quad (3-1)$$

The readers are also referred to the Appendix, which contains the rules for vector Gaussian distributions that are used in this derivation. Unlike Pearl, we consider normalizing constants for Gaussian distributions and show why they can be neglected.

### 3.1 Belief Update
Consider the general vector belief network in Figure 2. The links between the nodes correspond to
$$x = \sum_i F_i \cdot u_i + v$$
$$y_i = H_i x + w_i \quad (3-2)$$
The belief for node $X$ is
$$BEL(x) = f(x|e_x^+, e_x^-) = \alpha \cdot f(x|e_x^+) f(e_x^-|x) = \alpha \cdot \pi(x) \cdot \lambda(x)$$
$$\pi(x) = f(x|e_x^+) = \int_{U_1} \cdots \int_{U_n} f(x|e_x^+, u_1, \cdots, u_n) f(u_1, \cdots, u_n|e_x^+) du_1 \cdots du_n$$
$$= \int_{U_1} \cdots \int_{U_n} f(x|u_1, \cdots, u_n) \prod_{i=1}^n f(u_i|e_i^+) du_1 \cdots du_n$$
$$= \int_{U_1} \cdots \int_{U_n} N\left(x; Q, \sum_{i=1}^n B_i u_i\right) \prod_{i=1}^n N(u_i; P_i^*, \bar{u}_i^*) du_1 \cdots du_n$$
using Rule 3 from the appendix, we get
$$\pi(x) = N\left(x; \sum_{i=1}^n B_i P_{u_i} B_i^T + Q, \sum_{i=1}^n B_i \bar{u}_i^*\right) \quad (3-3)$$
$$= N[x; P_\pi, \bar{x}_\pi]$$

The variance $P_\pi$ can be interpreted as sum of the uncertainty in each of the $u_i$ (i.e., $\sum_{i=1}^{n} B_i P_{u_i} B_i^T$) and the uncertainty in the relationship between x and $u_i$ (i.e., $Q$). Similarly, we now compute $\lambda(x)$. Let $\Phi$ denote the set of child nodes of X, let $\Omega = \{j \in \Phi | e_j^- \neq \emptyset\}$, and let $m$ be the number of elements in $\Omega$. Next, we relabel the child nodes so that nodes $Y_1$ through $Y_m$ correspond to the nodes with $e_j^- \neq \emptyset$. If m = 0 then $\lambda(x) \stackrel{\Delta}{=} 1$. If m = 1 then $\lambda(x) \stackrel{\Delta}{=} \lambda_{Y_m}(x)$. For $m \geq 2$, we compute $\lambda(x)$ as follows

$$\lambda(x) = f(e_x^-|x) = f(e_1^-, e_2^-, \ldots, e_m^-|x) = \prod_j f(e_j^-|x) = \prod_j \lambda_j(x)$$

$$\lambda(x) = \prod_j N(H_j x; R_j, \bar{y}_j)$$

$$= a \cdot N\left(x; \left[\sum_j H_j^T R_j^{-1} H_j\right]^{-1}, \left[\sum_j H_j^T R_j^{-1} H_j\right]^{-1} \left[\sum_j H_j^T R_j^{-1} \bar{y}_j\right]\right)$$

$$= a \cdot N(x; P_\lambda, \bar{x}_\lambda)$$

where again $a$ is some constant (refer to Rule 5, appendix) the exact form of which is not important. As we will soon see it cancels out during the belief update process. To update $x$ we do not require $\left[\sum_j H_j^T R_j^{-1} H_j\right]^{-1}$ to be nonsingular. We also define a transformed state vector $z_\lambda \stackrel{\Delta}{=} P_\lambda^{-1} x_\lambda$ and thus $\bar{z}_\lambda = P_\lambda^{-1} \bar{x}_\lambda$. It is important to note that $P_\lambda^{-1}$ is really the covariance of the information state vector z, $P_\lambda^{-1} x_\lambda$ where $P_\lambda^{-1} = \sum_j H_j^T R_j^{-1} H_j$ and $P_\lambda^{-1} \bar{x}_\lambda = \sum_j H_j^T R_j^{-1} \bar{y}_j$. Hence,

$N(x; P_\lambda, \bar{x}_\lambda) = a_1 \cdot N(z; P_\lambda^{-1}, \bar{z})$, where again $a_1$ is some constant. This form of inference is particularly useful in deriving the decentralized version of the Kalman filter.
If $e_j^- = 0$ then $\lambda(x) \stackrel{\Delta}{=} 1$.
Combining these two results, we obtain
$BEL(x) = f(x|e_x^+, e_x^-)$

$$= \frac{f(e_x^-|x, e_x^+) \cdot f(x|e_x^+)}{\int_x f(e_x^-|x, e_x^+) \cdot f(x|e_x^+) \cdot dx} \quad (3\text{-}4)$$

$$= \frac{a \cdot N(x; P_\pi, \bar{x}_\pi) N(x; P_\lambda, \bar{x}_\lambda)}{a \cdot \int_x N(x; P_\pi, \bar{x}_\pi) N(x; P_\lambda, \bar{x}_\lambda) dx}$$

$$= \frac{N\left(x; [P_\pi^{-1} + P_\lambda^{-1}]^{-1}, [P_\pi^{-1} + P_\lambda^{-1}]^{-1} [P_\pi^{-1} \bar{x}_\pi + P_\lambda^{-1} \bar{x}_\lambda]\right) \cdot N(\bar{x}_\pi, P_\pi + P_\lambda, \bar{x}_\lambda)}{N(\bar{x}_\pi, P_\pi + P_\lambda, \bar{x}_\lambda)}$$

$$= N\left(x; [P_\pi^{-1} + P_\lambda^{-1}]^{-1}, [P_\pi^{-1} + P_\lambda^{-1}]^{-1} [P_\pi^{-1} \bar{x}_\pi + P_\lambda^{-1} \bar{x}_\lambda]\right)$$

$$= N\left(x; P_\pi - P_\pi [P_\pi + P_\lambda]^{-1} P_\pi, \bar{x}_\pi + P_\pi [P_\pi + P_\lambda]^{-1} (\bar{x}_\lambda - \bar{x}_\pi)\right)$$

As can be seen all constants associated with messages $\lambda(x)$ and $\pi(x)$ cancel out. Hence, the exact form of these constants is not important. Therefore, in the remaining part of this paper, we will neglect the constants.

To prescribe how the influence of new information will spread through the network, we need to specify how a typical node, say X, will compute its outgoing messages $\lambda_X(u_i)$, $i = 1, \ldots, n$, and $\pi_{Y_j}(x)$, $j = 1, \ldots, m$, from the incoming messages $\lambda_{Y_j}(x)$, $j = 1, \ldots, m$ and $\pi_X(u_i)$, $i = 1, \ldots, n$, which is done next.

### 3.2 Top Down Propagation: Message to Children
Consider the message $\pi_{Y_j}(x)$, which node X sends to its jth child $Y_j$ (j=1,2,..,m). We note that it is conditioned on all data except a subset $e_j^-$ of variables that connect to X via $Y_j$. Therefore,

$$\pi_{Y_j}(x) = f(x|e - e_j^-) = BEL(x|e_j^- = \phi)$$

$$= N(x; P_{Y_j}^*, \bar{x}_{Y_j}^*)$$

$$\pi_{Y_j}(x_j) = N(y_j; H_j P_{Y_j}^* H_j^T + R_i, H_j^T \bar{x}_{Y_j}^*)$$

So, $\pi_{Y_j}(x)$ can be computed by the method of the last section with the assumption that $\lambda_{Y_j}(x) = 1$. Hence,

$$P_{Y_j}^* = \left[P_\pi^{-1} + \sum_{k \neq j} H^T R_k^{-1} H\right]^{-1}$$

$$= P_\pi - P_\pi \left[P_\pi + \sum_{k \neq j} H^T R_k^{-1} H\right]^{-1} P_\pi$$

$$\bar{x}_{Y_j}^* = \left[P_\pi^{-1} + \sum_{k \neq j} H^T R_k^{-1} H\right]^{-1} \left[P_\pi^{-1} \bar{x}_\pi + \sum_{k \neq j} H^T R_k^{-1} \bar{y}_k\right]$$

$$= \bar{x}_\pi + \left[\sum_{k \neq j} H^T R_k^{-1} H\right] \left[P_\pi + \sum_{k \neq j} H^T R_k^{-1} H\right]^{-1} \left(\sum_{k \neq j} H^T R_k^{-1} \bar{y}_k - \bar{x}_\pi\right)$$

### 3.3 Bottom Up Propagation: Message to Parents
Consider the message $\lambda_X(u_i)$, which node X sends to its ith parent $U_i$. We divide the evidence e into its disjoint components $e_i^+$, i=1,..,n and $e_j^-$, j=1,..,m, and condition $\lambda_X(u_i)$ on all parents of X. For notational convenience we temporarily denote $U_i$ by U and $B_i$ by $B$, and let the other parents be indexed by $k$, ranging from $I$ to some $n$:

$$\lambda_X(u) = f(e - e_U^+|u)$$

$$= \int \cdots \int \int f(e_1^+, \ldots, e_n^+, e_1^-, \ldots, e_m^-|u_1, \ldots, u_n, x, u)$$

$$\cdot f(u_1, \ldots, u_n, x|u) dx \cdot du_1 \cdots du_n$$

Consider the first distribution in the integrand

$$f(\mathbf{e}_1^+,\ldots,\mathbf{e}_n^+,\mathbf{e}_1^-,\ldots,\mathbf{e}_m^-|u_1,\ldots,u_n,x,u)$$
$$= f(\mathbf{e}_1^-,\ldots,\mathbf{e}_m^-|x)\cdot f(\mathbf{e}_1^+,\ldots,\mathbf{e}_n^+|u_1,\ldots,u_n,x,u)$$
$$= \prod_j \lambda_{Y_j}(x)\cdot \prod_k f(\mathbf{e}_k^+|u_k)$$
$$= \lambda(x)\prod_k \frac{f(u_k|\mathbf{e}_k^+)f(\mathbf{e}_k^+)}{\int_{\mathbf{e}_k^+} f(u_k|\mathbf{e}_k^+)f(\mathbf{e}_k^+)d\mathbf{e}_k^+}$$

Next, consider the second distribution in the integrand
$$f(u_1,\ldots,u_n,x|u) = f(x|u,u_1,\ldots,u_n)f(u_1,\ldots,u_n)$$
$$= f(x|u,u_1,\ldots,u_n)\prod_k f(u_k)$$
$$= f(x|u,u_1,\ldots,u_n)\prod_k \int_{\mathbf{e}_k^+} f(u_k|\mathbf{e}_k^+)f(\mathbf{e}_k^+)d\mathbf{e}_k^+$$

Substituting these two distributions
$$\lambda_x(u) = \int_{u_1}\cdots\int_{u_n}\int_x \lambda(x)\prod_k f(u_k|\mathbf{e}_k^+)f(\mathbf{e}_k^+)f(x|u,u_1,\ldots,u_n)dx\cdot du_1\cdots du_n$$
$$= \int_{u_1}\cdots\int_{u_n}\int_x \lambda(x)\prod_k \pi_x(u_k)f(x|u,u_1,\ldots,u_n)dx\cdot du_1\cdots du_n$$

where
$$\lambda(x) = N(x;P_\lambda,\bar{x}_\lambda) = a\cdot N(z;P_\lambda^{-1},\bar{z}),$$
$$\pi_x(u_k) = N(u_k;P_{u_k}^+,\bar{u}_k),\ f(x|u,u_1,\ldots,u_n) = N\left(x;Q,Bu+\sum_k B_k u_k\right)$$

Using the properties for vector normal distributions
$$\lambda_x(u) = \int_{u_1}\cdots\int_{u_n}\int_x N(x;P_\lambda,\bar{x}_\lambda)\prod_k N(u_k;P_{u_k}^+,\bar{u}_k)$$
$$\cdot N\left(x;Q,Bu+\sum_k B_k u_k\right)dx\cdot du_1\cdots du_n$$

Integrate with respect to x
$$= \int_{u_1}\cdots\int_{u_n}\prod_k N(u_k;P_{u_k}^+,\bar{u}_k)N\left(\bar{x}_\lambda;P_\lambda+Q,Bu+\sum_k B_k u_k\right)du_1\cdots du_n$$

$$\lambda_x(u) = N\left(Bu;P_\lambda+Q+\sum_k B_k P_{u_k}^+ B_k^T,\bar{x}_\lambda-\sum_k B_k \bar{u}_k\right)$$

$$= N\left(\begin{array}{l} u;P_x(u) = \left[B^T\left(P_\lambda+Q+\sum_k B_k P_{u_k}^+ B_k^T\right)^{-1}B\right]^{-1}, \\ P_x(u)\left[B^T\left(P_\lambda+Q+\sum_k B_k P_{u_k}^+ B_k^T\right)^{-1}\left(\bar{x}_\lambda-\sum_k B_k \bar{u}_k\right)\right]\end{array}\right)$$

Therefore, for the *i*th parent, $U_i$ we have
$$\lambda_x(u_i) = N(u_i;P_x^-(u_i),\bar{x}_x^-(u_i))$$

$$N\left(\begin{array}{l} u_i;\left[B_i^T\left(P_\lambda+Q+\sum_{k\ne i} B_k P_{u_k}^+ B_k^T\right)^{-1}B_i\right]^{-1}, \\ P_x^-(u_i)\left[B_i^T\left(P_\lambda+Q+\sum_{k\ne i} B_k P_{u_k}^+ B_k^T\right)^{-1}\left(\bar{x}_\lambda-\sum_{k\ne i} B_k \bar{u}_k\right)\right]\end{array}\right)$$

where

$$P_\lambda = \left[\sum_j H_j^T R_j^{-1} H_j\right]^{-1};\quad P_\lambda^{-1} = \sum_j H_j^T R_j^{-1} H_j$$
$$\bar{x}_\lambda = \left[\sum_j H_j^T R_j^{-1} H_j\right]^{-1}\left[\sum_j H_j^T R_j^{-1}\bar{y}_j\right];\quad P_\lambda^{-1}\bar{x}_\lambda = \sum_j H_j^T R_j^{-1}\bar{y}_j$$

Appendix (Section 7.2) contains an alternate form of message to the parents and belief update.

### 3.4 Predictive Estimation: No Evidence in Children Nodes

If $\lambda(x) = 1$ — there is no evidence from the children — we have
$$\lambda_x(u) = \int_{u_1}\cdots\int_{u_n}\int_x \prod_k N(u_k;P_{u_k}^+,\bar{u}_k)$$
$$\cdot N\left(x;Q,B\bar{u}+\sum_k B_k \bar{u}_k\right)dx\cdot du_1\cdots du_n$$
$$= \int_{u_1}\cdots\int_{u_n}\prod_k N(u_k;P_{u_k}^+,\bar{u}_k)du_1\cdots du_n$$
$$= 1$$

which implies that evidence gathered at a node does not affect any of its spouses until their common child node obtains evidence. This reflects the *d*-separation condition and matches our intuition regarding multiple causes.

### 3.5 Boundary Conditions
The boundary conditions for vector Gaussian continuous networks are established as follows:

1. If X is a root node – a node with no parents – that has not been instantiated, then we set $\pi(x)$ equal to the prior density function.

2. If X is a leaf node – a node with no children – that has not been instantiated, then we set $\lambda(x) = 1$. This implies that $Bel(x) = \pi(x)$.

3. If X is an evidence node, say $X = \bar{x}$, then we set $\lambda(x) = \delta(x-\bar{x}) = N(x;0,\bar{x})$ regardless of the incoming $\lambda$-messages. This implies that $Bel(x) = N(x;0,\bar{x})$ as expected. Furthermore, for each *j*, $\pi_{Y_j}(x) = N(x;0,\bar{x})$ is the message that node X sends to its children (in this case each child gets the same message).

## 4 Inference Using Topology Transformation
Another method for doing inference in a belief network and for decision-making in influence diagrams – belief network with decision nodes – consists of eliminating nodes and transforming the diagram through a series of transformations (Olmsted, 1984; Shachter, 1986; Rege and Agogino, 1988). For vector Gaussian continuous networks we require two basic transformations, which are developed next.

## 4.1 Parent Node Removal

The process of removing a parent node, i.e., node propagation in the direction of the arrow, corresponds to taking the expectation with respect to that parent variable. We make use of the generalized sum rule

$$\Pr(X|\xi) = \int_y \Pr(X,Y|\xi) dy \quad (4\text{-}1)$$

Figure 3 shows the change in topology of the vector Gaussian belief network due to this transformation. Let,

$$x = Fu + \sum_i F_i u_i + v \quad (4\text{-}2)$$

correspond to the relationship between $x$ and its parent nodes. Let, $Q$ be the covariance of the noise in the relationship. Then after the transformation we have the relationship

$$x = F\overline{u} + \sum_j F_j u_j + v' \quad (4\text{-}3)$$

where the distribution for $u$ was given by $N(u; P_u, \overline{u})$, the covariance of noise $v'$ is given by $Q + FP_u F^T$.

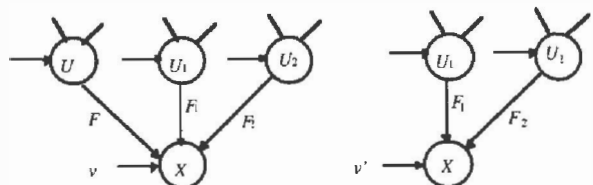

*Figure 3: Topology transformation after parent node removal.*

## 4.2 Arc Reversal and Propagation

The process of arc reversal, and node propagation corresponds to applying Bayes' rule, which for Gaussian random variables (for a proof see Bar-Shalom and Li, 1993; Pages 43-44), is given by

$$\textit{Mean: } E[x|y] \overset{\Delta}{=} \hat{x} = \overline{x} + P_{xy} P_{yy}^{-1}(y - \overline{y})$$

$$\textit{Variance: } Cov[x|y] \overset{\Delta}{=} P_{x|y} = T_{xx}^{-1} = P_{xx} - P_{xy} P_{yy}^{-1} P_{yx} \quad (4\text{-}4)$$

which is the conditional probability of x given y. Here, y corresponds to the value of the variables in the child node, which is to be removed from the graph. After the application of Equation 4-4 the node is removed from the graph as shown in Figure 4.

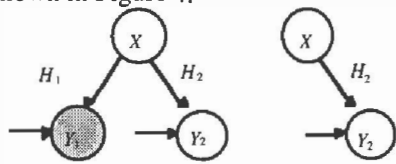

*Figure 4: Topology transformation after arc reversal and propagation.*

## 5 Example: Kalman Filter

Here, we illustrate the potential of these networks for developing Bayesian algorithms by deriving the decentralized form of the Kalman filter. The centralized form can be derived using the rules for topology transformation. This derivation can be found in Alag (1996; Chapter 3) along with a detailed numerical example on carrying out inference in these networks (Section 2.8). The Kalman filter is a form of optimal estimation in the statistical sense characterized by recursive evaluation, an internal model of the dynamics of the system being estimated, and a dynamic weighting of incoming evidence with ongoing expectation that produces estimates of the state of the observed system (Bar-Shalom and Li, 1993).

### 5.1 Problem Statement

Consider a discrete time linear stochastic system described by the following vector difference equation

$$x(k+1) = F(k)x(k) + G(k)u(k) + v(k) \quad k = 0,1,..$$

where vector $x$ of dimension $n_z$ is the state of the system. $F$ is the state transition matrix, $G$ is the discrete time gain through which the input, assumed to be constant over a sampling period, enters the system. We have introduced time through the index $k$. $u(k)$ is an $n_z$-dimensional *known input vector*, and $v(k)$, $k=0,1,..$, is the sequence of zero-mean white Gaussian *process noise* (also $n_z$ dimension vector) with covariance

$$E[v(k)v(k)^T] = Q(k)$$

The system is being monitored by a group of sensor with the following sensor model

$$z(k) = H(k)x(k) + w(k) \quad k = 1,...$$

where $H$ is the measurement matrix, $w(k)$ the sequence of zero-mean white Gaussian measurement noise with covariance

$$E[w(k)w(k)^T] = R(k)$$

The observations could be some simple subset of the parameters, or the complete parameter vector. It is assumed that the state evolution model: matrices F(k), G(k), Q(k), as well as the sensor model: matrices H(k), R(k) are known.

The initial state of the random variable $x(0)$ is generally unknown and the probability density function is modeled as a normal distribution with a known mean and variance. The two noise sequences and the initial state are assumed mutually independent.

### 5.2 Graphical Representation

We include time in our representation using dynamic belief network framework (see for e.g., Russell and Norvig, 1995). As shown in Figure 5 there are three kinds of nodes: state node $x$, the control nodes $u$, the sensor nodes $z$. The arc from $x(k)$ to $x(k+1)$ corresponds to projection in time – the state evolution model.

26   Alag and Agogino

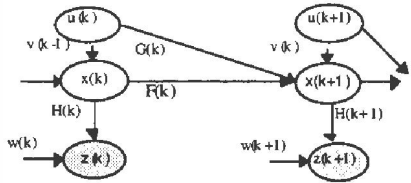

*Figure 5: Vector Gaussian dynamic belief network for the Kalman filter.*

### 5.3 Inference: Decentralized Kalman Filter

We shall follow the three step methodology for doing inference in dynamic belief networks (Russell and Norvig, 1995). Consider the network at time $k$. Let our estimate for the vector variables $x(k)$ after using all the evidence at time k be $N\left(x(k), P(k|k), \hat{x}(k|k)\right)$.

*(1) Prediction*

Figure 6 shows the belief network for this phase. Note, that $u(k)$ is a deterministic node and it only shifts the mean. It has no effect on the variance.

$$\pi(x(k+1)) = N\left(x(k+1, P(k+1|k), \hat{x}(k+1|k))\right)$$

$$= N\left(x(k+1); F(k)P(k|k)F(k)^T + Q, F(k)\hat{x}(k|k) + G(k)u(k)\right)$$

$$\lambda(x(k+1)) = 1$$

$$Bel(x(k+1)) = \pi(x(k+1))$$

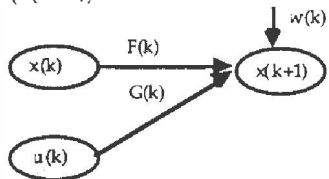

*Figure 6: The belief network at the beginning of the prediction stage.*

*(2) Roll Up*

The slice at time k is removed and the prior distribution for x(k+1) is $N\left(x(k+1); P(k+1|k), \hat{x}(k+1|k)\right)$.

*(3) Estimation*

Next as shown in Figure 7 the new evidence in the form of sensor readings is used to update our belief for $x(k+1)$.

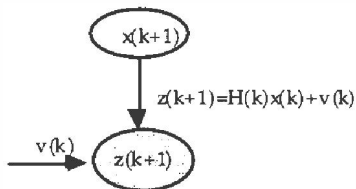

*Figure 7: The belief network during the estimation stage.*

$$\pi(x) = N\left(x(k+1), P_\pi, \bar{x}_\pi\right) = N\left(x(k+1), P(k+1|k), \hat{x}(k+1|k)\right)$$

$$\lambda(x) = N\left(x(k+1), P_\lambda, \bar{x}_\lambda\right) =$$

$$N\left(\begin{array}{l} x(k+1), \left[H(k+1)^T R^{-1}(k+1)H(k+1)\right]^{-1}, \\ \left[H(k+1)^T R^{-1}(k+1)H(k+1)\right]^{-1}\left[H(k+1)^T R^{-1}z(k+1)\right] \end{array}\right)$$

$$Bel(x) = N\left(x(k+1), P(k+1|k+1), \hat{x}(k+1|k+1)\right)$$

$$= N\left(\begin{array}{l} x(k+1); \left[P^{-1}(k+1|k) + H(k+1)^T R^{-1}(k+1)H(k+1)\right]^{-1}, \\ P^{-1}(k+1|k+1)\left(P^{-1}(k+1|k)\hat{x}(k+1|k) + H(k+1)^T R^{-1}z(k+1)\right) \end{array}\right)$$

which is the decentralized form of the Kalman filter. For more details see Alag (1996; Chapter 3).

## 6 Conclusions

In this paper we have extended Gaussian networks to its vector form, where a node represents multivariate Gaussian distribution. These networks have been developed to represent uncertainty inherent in the estimation process for dynamic systems. We have developed rules for inference in these networks using both the method of message propagation and topology transformation. The two different methods for inference lead to development of algorithms which can be implemented with either multi-processors (in a parallel or a decentralized manner) or with a single processor (centralized manner). Using the network structure and the rules for inference developed here, important algorithms such as the Kalman filter, the probabilistic data association filter, interacting multiple model algorithm can be represented and derived. Readers are referred to Alag (1996) for the application of this representation and inference rules to develop algorithms for monitoring and diagnosis of complex systems and their application to extant practical systems, namely power plant monitoring and automated vehicles. This paper unifies concepts from artificial intelligence and modern control theory.

**Acknowledgments**

The work was sponsored by Partners for Advanced Transit Highways (PATH) through MOU-132 and MOU-157.

Research, Advanced Technology Division, Microsoft Corporation, WA.

# 7 Appendix
## 7.1 Rules for Vector Gaussian Distribution[4]

1. $N(x; P, \bar{x}) = |2\pi P|^{-1/2} \exp\left(-\frac{1}{2}(x-\bar{x})^T P^{-1}(x-\bar{x})\right)$

2. $N(x; P, \bar{x}) = N(\bar{x}, P, x)$

3. $N(y = Ax + B; P_y, \bar{y}) = a \cdot N\begin{pmatrix} x; P_x = [A^T P_y^{-1} A]^{-1}, \\ P_x [A^T P_y^{-1}(\bar{y} - B)] \end{pmatrix}$

where $a$ is some constant.

4. $N(x; P_1, \bar{x}_1) \cdot N(x; P_2, \bar{x}_2) = a \cdot N\begin{pmatrix} x; [P_1^{-1} + P_2^{-1}]^{-1}, \\ [P_1^{-1} + P_2^{-1}]^{-1}[P_1^{-1}\bar{x}_1 + P_2^{-1}\bar{x}_2] \end{pmatrix}$

where the constant $a$ is given by
$a = N(\bar{x}_1, P_1 + P_2, \bar{x}_2)$

5. $N(x; P_1, \bar{x}_1) \cdot N(x; P_2, \bar{x}_2) = a \cdot N\begin{pmatrix} x; P_2 - P_2[P_1 + P_2]^{-1} P_2, \\ \bar{x}_2 + P_2[P_1 + P_2]^{-1}(\bar{x}_1 - \bar{x}_2) \end{pmatrix}$

here again a is some constant.

6. $\prod_i N(Ax; P_i, \bar{y}_i) = a \cdot N\begin{pmatrix} x; \left[\sum_i A^T P_i^{-1} A\right]^{-1}, \\ \left[\sum_i A^T P_i^{-1} A\right]^{-1}\left[\sum_i A^T P_i^{-1} \bar{y}_i\right] \end{pmatrix}$

where again a is some constant.

7. $\int N(y; P_1, \bar{x}_1) N(y; P_2, x) dy = N(x; P_1 + P_2, \bar{x}_1)$

8. $\int_y N(x; P_x, y) N(y; P_y, \bar{y}) dy = N[x; P_x + P_y, \bar{y}]$

---

[4] Refer to Alag (1996; Chapter 2) for detailed proofs.

9. $\int_{u_1} \cdots \int_{u_n} \prod_{i=1}^{n} N(u_i, P_i, \bar{u}_i) \cdot N\left(\sum_j B_j u_j, Q, x\right) du_1 \cdots du_n$

$= N\left(x; Q + \sum_j B_j P_j B_j^T, \sum_j B_j \bar{u}_j\right)$

## 7.2 Alternate Forms for Message to Parents

We can simplify the belief update process for the parent node $u_i$ (continued from Section 3.3)

$P_\pi(u_i) = N(u_i, P_{u_i}, \bar{u}_i) = N(B_i u_i; B_i P_{u_i} B_i^T, B_i \bar{u}_i)$

$P_\lambda(u_i) = N\left(B_i u_i; P_\lambda + Q + \sum_{k \neq i} B_k P_{u_k} B_u^T, \bar{x}_\lambda - \sum_{k \neq i} B_k \bar{u}_k\right)$

Combining the above two equations to update the belief in $u_i$

$N(B_i u_i; B_i P_{u_i}^{new} B_i^T, B_i \bar{u}_i^{new})$

$B_i P_{u_i}^{new} B_i^T = (B_i P_{u_i} B_i^T) - (B_i P_{u_i} B_i^T)\left[P_\lambda + Q + \sum_k B_k P_{u_k} B_u^T\right]^{-1}(B_i P_{u_i} B_i^T)$

$B_i \bar{u}_i^{new} = B_i \bar{u}_i + (B_i P_{u_i} B_i^T)\left[P_\lambda + Q + \sum_e B_k P_{u_k} B_u^T\right]^{-1}\left(\bar{x}_\lambda - \sum_k B_k \bar{u}_k\right)$

Therefore,

$N(u_i; P_{u_i}^{new}, \bar{u}_i^{new})$

$= N\begin{pmatrix} u_i; P_{u_i} - P_{u_i} B_i^T \left[P_\lambda + Q + \sum_k B_k P_{u_k} B_u^T\right]^{-1} B_i P_{u_i}, \\ \bar{u}_i + P_{u_i} B_i^T \left[P_\lambda + Q + \sum_k B_k P_{u_k} B_u^T\right]^{-1}\left(\bar{x}_\lambda - \sum_k B_k \bar{u}_k\right) \end{pmatrix}$

when the above formula is used we require $P_\lambda = \left[\sum_j H_j^T R_j^{-1} H_j\right]^{-1}$ to exist. We can remove this requirement by noting

$P_\lambda^{-1} = \sum_{j=1}^{m} H_j^T R_j^{-1} H_j \stackrel{\Delta}{=} H^T R^{-1} H$

$H = [H_1^T, \cdots, H_m^T]^T \quad R = blockdiag\{R_1, \cdots, R_m\}$

$\therefore R = HP_\lambda H^T \quad P_\lambda^{-1} = H^T[HP_\lambda H^T]^{-1} H$

where $m$ is the number of children nodes of $X$ through which there is evidence. But

$\left[P_\lambda + Q + \sum_k B_k P_{u_k} B_u^T\right]^{-1} = H^T\left[H\left(Q + \sum_k B_k P_{u_k} B_u^T\right)H^T + R\right]^{-1} H$

Further,

$P_\lambda^{-1} \bar{x}_\lambda = \sum_{j=1}^{m} H^T R_j^{-1} \bar{y}_j = H^T R^{-1} \bar{y}$

$H^T R^{-1} H \bar{x}_\lambda = H^T R^{-1} \bar{y}$

$\therefore H \bar{x}_\lambda = \bar{y}$

$N(u_i; P_{u_i}^{new}, \bar{u}_i^{new}) = N\begin{pmatrix} u_i; P_{u_i} - P_{u_i} B_i^T H^T\left[H\left(Q + \sum_k B_k P_{u_k} B_u^T\right)H^T + R\right]^{-1} HB_i P_{u_i}, \\ \bar{u}_i + P_{u_i} B_i^T H^T\left[H\left(Q + \sum_k B_k P_{u_k} B_u^T\right)H^T + R\right]^{-1}\left(\bar{y} - H\sum_k B_k \bar{u}_k\right) \end{pmatrix}$

is an alternate form for updating the beliefs.